\documentclass[10pt,twocolumn,letterpaper]{article}

\usepackage{iccv}
\usepackage{times}
\usepackage{epsfig}
\usepackage{graphicx}
\usepackage{amsmath}
\usepackage{amssymb}
\usepackage{cite}
\usepackage{booktabs}

\usepackage[pagebackref=true,breaklinks=true,letterpaper=true,colorlinks,bookmarks=false]{hyperref}

\iccvfinalcopy 

\def\iccvPaperID{1788} 

\def\figvspace{{\vspace{-3mm}}}
\def\eqnvspace{{\vspace{-3mm}}}

\newcommand{\unmarkedfntext}[1]{
	\begingroup
	\renewcommand\thefootnote{}\footnote{#1}
	\addtocounter{footnote}{-1}
	\endgroup
}

\newcommand{\Paragraph}[1]{\vspace{-0mm} \noindent \textbf{#1} \hspace{0mm}}

\ificcvfinal\fi
\begin{document}

\title{Towards Interpretable Face Recognition}

\author{Bangjie Yin$^{1*}$
\hspace{0.08in} Luan Tran$^{1*}$
\hspace{0.08in} Haoxiang Li$^{2\dagger}$
\hspace{0.08in} Xiaohui Shen$^{3\dagger}$
\hspace{0.08in} Xiaoming Liu$^{1}$
\vspace{1mm} \\
\hspace{0.15in} $^{1}$Michigan State University
\hspace{0.15in} $^{2}$Wormpex AI Research
\hspace{0.15in} $^{3}$ByteDance AI Lab\\
}

\maketitle

\begin{abstract}
   Deep CNNs have been pushing the frontier of visual recognition over past years. Besides recognition accuracy, strong demands in understanding deep CNNs in the research community motivate developments of tools to dissect pre-trained models to visualize how they make predictions. Recent works further push the interpretability in the network learning stage to learn more meaningful representations. In this work, focusing on a specific area of visual recognition, we report our efforts towards interpretable face recognition. We propose a spatial activation diversity loss to learn more structured face representations. By leveraging the structure, we further design a feature activation diversity loss to push the interpretable representations to be discriminative and robust to occlusions. We demonstrate on three face recognition benchmarks that our proposed method is able to achieve the state-of-art face recognition accuracy with easily interpretable face representations.
\end{abstract}

\unmarkedfntext{{$^*$} Denotes equal contribution by the authors. $^\dagger$ Dr. Li and Dr. Shen contributed to this work while employed by Adobe Inc. Project page is at \url{http://cvlab.cse.msu.edu/project-interpret-FR}}

\section{Introduction}

In the era of deep learning, one major focus in the research community has been on designing network architectures and objective functions towards discriminative feature learning~\cite{he2016deep,iandola2014densenet,lin2017focal,wen2016discriminative,liu2017learning, tran2017missing}. 
Meanwhile, given its superior even surpassing-human recognition accuracy~\cite{he2015delving,lu2015surpassing}, there is a strong demand from both researchers and general audiences to interpret its successes and failures~\cite{goodfellow2014explaining,olah2018building}, to understand, improve, and trust its decisions. 
Increased interests in visualizing CNNs lead to a set of useful tools to dissect their prediction paths to identify the important visual cues~\cite{olah2018building}. 
While it is interesting to see the visual evidences for predictions from pre-trained models, what's more interesting is to guide the learning towards better interpretability.

CNNs trained towards discriminative classification may learn filters with wide-spreading attentions -- usually hard to interpret for human. Prior work even empirically demonstrate models and human attend to different image areas in visual understanding~\cite{das2017human}. 
Without design to harness interpretability, even when filters are observed to actively respond to certain local structure across several images, there is nothing preventing them to simultaneously capture a different structure; and the same structure may activate other filters too. 
One potential solution to address this issue is to provide annotations to learn locally activated filters and construct a structured representation from bottom-up. 
However, in practice, this is rarely feasible. 
Manual annotations are expensive to collect, difficult to define in certain tasks, and sub-optimal compared with end-to-end learned filters.

\begin{figure}[t]
    \centering
    \includegraphics[width=\linewidth]{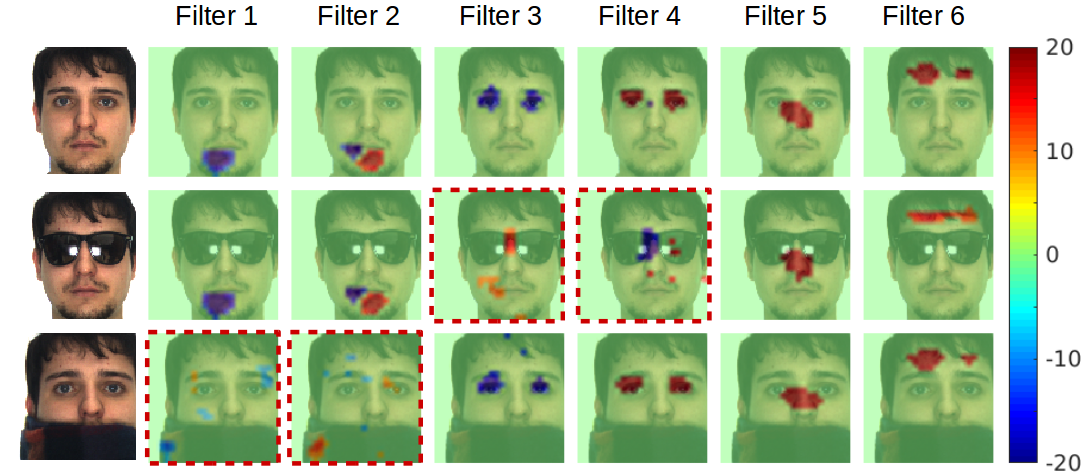}
    \caption{\small An example on the behaviors of an interpretable face recognition system: left most column is three faces of the same identity and right six columns are filter responses from six filters; each filter captures a clear and consistent semantic face part, e.g., eyes, nose, and jaw; heavy occlusions, eyeglass or scarf, alternate responses of corresponding filters and make the responses being more scattered, as shown in red bounding boxes.}
    \label{fig:top}\figvspace
\end{figure}

A desirable solution would keep the end-to-end training pipeline intact and encourage the interpretability with a model-agnostic design. However, in the recent interpretable CNNs~\cite{zhang2017interpretable}, where filters are trained to represent object parts to make the network representation interpretable, they observe degraded recognition accuracy after introducing interpretability. While the work is seminal and inspiring, this drawback largely limits its practical applicability.


In this paper, we study face recognition and strive to learn an interpretable face representation (Fig.~\ref{fig:top}). We define interpretability in this way that when each dimension of the representation is able to 
represent a face structure or a face part, the face representation is of higher interpretability.
Although the concept of part-based representations has been around~\cite{li2001learning,felzenszwalb2008discriminatively,berg2013poof,li2017probabilistic}, prior methods are not easily applicable to deep CNNs. 
Especially in face recognition, as far as we know, this problem is rarely addressed in the literature. 


In our method, the filters are learned end-to-end from data and constrained to be locally activated with the proposed spatial activation diversity loss. 
We further introduce a feature activation diversity loss to better align filter responses across faces and encourage filters to capture more discriminative visual cues for face recognition, especially occluded face recognition. 
Compared with the interpretable CNNs from Zhang et al.~\cite{zhang2017interpretable}, our final face representation does not compromise recognition accuracy, instead it achieves improved performance as well as enhanced robustness to occlusion. 
We empirically evaluate our method on three face recognition benchmarks with detailed ablation studies on the proposed objective functions.


To summarize, our contributions in this paper are in three-fold: 
1) we propose a spatial activation diversity loss to encourage learning interpretable face representations; 
2) we introduce a feature activation diversity loss to enhance discrimination and robustness to occlusions, which promotes the practical value of interpretability; 
3) we demonstrate superior interpretability, while achieving improved or similar face recognition performance on three face recognition benchmarks, compared to base CNN architectures.

\section{Related Work}

\Paragraph{Interpretable Representation Learning}
Understanding the visual recognition has a long history in computer vision~\cite{mahendran2016visualizing,sudderth2005learning,juneja2013blocks,singh2012unsupervised,parikh2011human}. 
In early days when most models use hand-craft features, a number of research focused on how to interpret the predictions. 
Back then visual cues include image patches~\cite{juneja2013blocks}, object colors~\cite{szegedy2013properties}, body parts~\cite{yao2011human}, face parts~\cite{li2017probabilistic}, or middle-level representations~\cite{singh2012unsupervised} contingent on the tasks. 
For example, Vondrick et al.~\cite{vondrick2013hoggles} develop the HOGgles to visualize HOG descriptors in object detection. 
Since features such as SIFT~\cite{lowe2004distinctive}, LBP~\cite{ahonen2006face} are extracted from image patches and serve as building blocks in the recognition pipeline, it was intuitive to describe the process from the level of patches. 
With the more complicated CNNs, it demands new tools to dissect its prediction. 
Early works include direct visualization of the filters~\cite{zeiler2014visualizing}, deconvolutional networks to reconstruct inputs from different layers~\cite{zeiler2011adaptive}, gradient-based methods to generate novel inputs that maximize certain neurons~\cite{nguyen2015deep}, and etc. 
Recent efforts along this line include CAM~\cite{zhou2016learning} which leverages the global max pooling to visualize dimensions of the representation and Grad-CAM~\cite{selvaraju2016grad} which relaxes the constraints on the network with a general framework to visualize any convolution filters. 
While our method can be related to visualization of CNNs and we leverage tools to visualize our learned filters, it is not the focus of this paper.

Visualization of CNNs is a good way to interpret the network but by itself it does not make the network more interpretable. 
Attention model~\cite{xu2015attention} has been used in image caption generation. 
By attention mechanism, their model can push the feature maps responding separately to each predicted caption word, which is seemingly close to our idea, but needs many labeled data for training. 
One recent work on learning a more meaningful representation is the interpretable CNNs~\cite{zhang2017interpretable}, where two losses regularize the training of late-stage convolutional filters: one to encourage each filter to encode a distinctive object part and another to push it to respond to only one local region. 
AnchorNet~\cite{novotny2017anchornet} adopts the similar idea to encourage orthogonality of filters and filter responses to keep each filter activated by a local and consistent structure. 
Our method generally extends the ideas in AnchorNet with new aspects for face recognition in designing our spatial activation diversity loss. Another line of research in learning interpretable representations is also referred to as feature disentangling, e.g., image synthesis/editing~\cite{chen2016infogan,shu2017neural}, face modeling~\cite{tran2018nonlinear,tran2019on,tran2019towards} and recognition~\cite{zhang2019gait,disentangling-features-in-3d-face-shapes-for-joint-face-reconstruction-and-recognition}. 
They intend to factorize the latent representation to describe the inputs from different aspects, of which the direction is largely diverged from our goal in this paper. 

\begin{figure*}
\center
\includegraphics[width=0.76\linewidth]{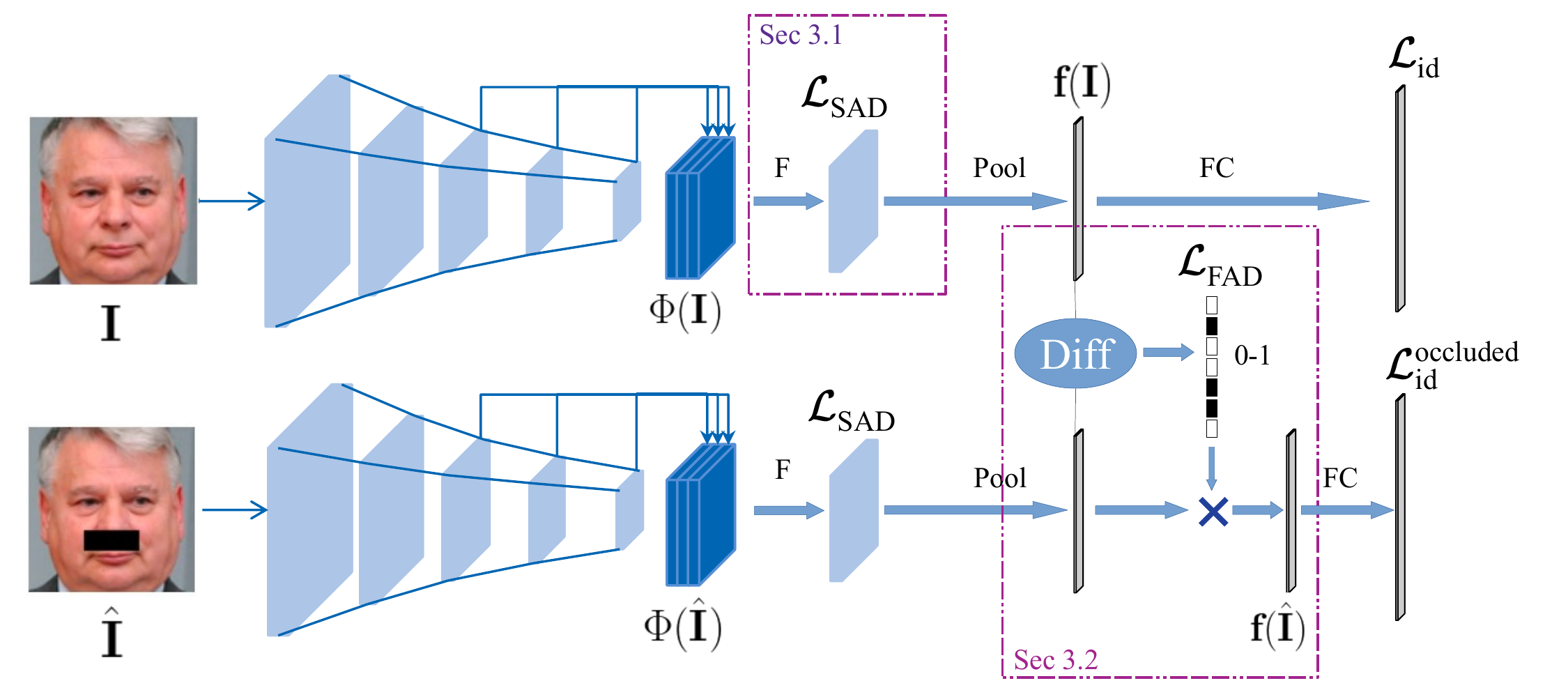}
\caption{\small The overall network architecture of the proposed framework. While the Spatial Activation Diversity (SAD) loss promotes structured feature responses, Feature Activation Diversity (FAD) loss enforces them to be insensitive to local changes (occlusions). }
\label{fig:network_arch}
\figvspace 
\end{figure*}

\Paragraph{Parts and Occlusion in Face Recognition}
As an extensively studied topic~\cite{learned2016labeled, OToole2018facespace,chen2002principle}, early face recognition works constructing meaningful representations mostly aim to improve the recognition accuracy. 
Some representations are composed from face parts.
The part-based models are either learned unsupervisedly from data~\cite{li2013probabilistic} or specified by manually annotated landmarks~\cite{cao2010face}. 
Besides local parts, face attributes are also interesting elements to build up representations. 
Kumar et al.~\cite{kumar2009attribute} encode a face image with scores from attribute classifiers and demonstrate improved performance before the deep learning era. 
In this paper, we propose to learn meaningful part-based face representations with a deep CNN, through 
the carefully designed losses. 
We demonstrate how to leverage the interpretable representation for occlusion-robust face recognition.
Prior methods addressing face pose variations~\cite{liu2003geometry,liu2006optimal,li2013probabilistic,cao2010face,tran2017disentangled,chai2007locally,towards-large-pose-face-frontalization-in-the-wild,multi-task-convolutional-neural-network-for-pose-invariant-face-recognition} can be related since pose changes may lead to self-occlusions. 
However, this work is interested in more explicit situations when faces are occluded by hand, sunglasses, and other objects. Interestingly, this specific aspect is rarely studied with CNNs. 
Cheng et~al.~\cite{cheng2015robust} propose to restore occluded faces with deep auto-encoder for improved recognition accuracy.  
Zhou et al.~\cite{zhou2015naive} argue that naively training a high capacity network with sufficient coverage in training data could achieve superior accuracy. 
In our experiment, we indeed observe improved recognition accuracy to occluded faces after augmenting training with synthetic occluded faces. 
However, with the proposed method, we can further improve robustness to occlusion without increasing network capacity, which highlights the merits of interpretable representation.

\Paragraph{Occlusion Handling with CNNs}
Different methods are proposed to handle occlusion with CNNs for robust object detection and recognition. 
Wang et~al.~\cite{AFastRcnn} learn an object detector by generating an occlusion mask for each object, which synthesizes harder samples for the adversarial network. 
In~\cite{HideAndSeek}, occlusion masks are utilized  to enforce the network to pay attention to different parts of the objects. 
Ge et al.~\cite{LLECNN} detect faces with heavy occlusions by proposing a masked face dataset and applying it to their proposed LLE-CNNs. 
In contrast, our method enforces constraints for the spreadness of  feature activations and guides the network to extract features from different face parts.

\section{Proposed Method}

Our network architecture in training is shown in Fig.~\ref{fig:network_arch}.  
From a high-level perspective, we construct a Siamese network with two branches sharing weights to learn face representations from two faces: one with synthetic occlusion and one without. 
We would like to learn a set of diverse filter $\mathbf{F}$, which applies on a hypercolumn (HC) descriptor $\Phi$, consisting of feature at multiple semantic levels. 
The proposed Spatial Activation Diversity (SAD) loss encourages the face representation to be structured with consistent semantic meaning. 
Softmax loss helps encode the identity information. 
The input to the lower network branch is a synthetic occluded version of the above input. 
The proposed Feature Activation Diversity (FAD) loss requires filters to be insensitive to the occluded part, hence more robust to occlusion.
At the same time, we mask out parts of the face representation sensitive to the occlusion and train to identify the input face solely based on the remaining elements. 
As a result, the filters respond to non-occluded parts are trained to capture more discriminative cues for identification. 

\subsection{Spatial Activation Diversity Loss}

Novotny et al.~\cite{novotny2017anchornet} propose a diversity loss for semantic matching by penalizing correlations among filters weights and their responses. 
While their idea is general enough to extend to face representation learning, in practice, their design is not directly applicable due to the prohibitively large number of identities (classes) in face recognition. 
Their approach also suffers from degradation in recognition accuracy. 
We first introduce their diversity loss and then describe our proposed modifications tailored to face recognition.

\Paragraph{Spatial Activation Diversity Loss} For each of $K$ class in the training set, Novotny et al.~\cite{novotny2017anchornet} propose to learn a set of diverse filters with discriminative power to distinguish an object of the category and background images. 
The filers $\mathbf{F}$ apply on a hypercolumn descriptor $\Phi(\mathbf{I})$, created by concatenating the filter responses of an image $\mathbf{I}$ at different convolutional  layers~\cite{hariharan2015hypercolumns}. 
This helps $\mathbf{F}$ to aggregate features at different semantic levels. 
The response map of this operation is denoted as $\psi(\mathbf{I}) = \mathbf{F} * \Phi(\mathbf{I})$.

The diversity constraint is implemented by two \emph{diversity losses} $\mathcal{L}_{\text{SAD}}^{\text{filter}}$ and $\mathcal{L}_{\text{SAD}}^{\text{response}}$, encouraging the orthogonality of the filters and of their responses, respectively. 
$\mathcal{L}_{\text{SAD}}^{\text{filter}}$ makes filters orthogonal by penalizing their correlations:
\begin{equation}
    \mathcal{L}_{\text{SAD}}^{\text{filter}} (\mathbf{F}) =  \sum_{i \neq j} 
    \left|
    \sum_p
    \frac{\langle \mathbf{F}_i^p, \mathbf{F}_j^p \rangle}{\|\mathbf{F}_i^p\|_F ~ \|\mathbf{F}_j^p\|_F } 
    \right|,
    \label{eq:div1}
\end{equation}
where $\mathbf{F}_i^p$ is the column of filter $\mathbf{F}_i$ at the spatial location $p$.
Note that orthogonal filters are likely to respond to different image structures, but this is not necessarily the case. Thus, the second term $\mathcal{L}_{\text{SAD}}^{\text{response}}$ is introduced to directly decorrelate the filters' \emph{response maps} $\psi_i(\mathbf{I})$:
\begin{equation}
\mathcal{L}_{\text{SAD}}^{\text{response}} (\mathbf{I};\Phi,\mathbf{F}) = \sum_{i \neq j}
  \left\lVert
  \frac{\langle \psi_i,  \psi_j\rangle} {\|\psi_i\|_F \|\psi_j\|_F}
  \right\rVert^2.
\label{eq:div2}
\end{equation}
This term is further regularized by using the smoothed response maps $\psi'(\mathbf{I}) \doteq g_\sigma * (\psi(\mathbf{I}))$ in place of $\psi(\mathbf{I})$ in $\mathcal{L}_{\text{SAD}}^{\text{response}}$ loss computing. Here the channel-wise Gaussian kernel $g_\sigma$ is applied to encourage filter responses to spread further apart by dilating their activations.

\Paragraph{Our Proposed Modifications} Novotny et al.~\cite{novotny2017anchornet} learn $K$ sets of filters, one for each of $K$ categories. The discrimination of the features are maintained by $K$ binary classification losses for each category vs.~background images. The discriminative loss is proposed to enhance (or suppress) the maximum value in the response maps $\psi_i$ for the positive  (or negative) class. 
In ~\cite{novotny2017anchornet}, the final feature representation $\mathbf{f}$ is obtained via global max-pooling operation on $\psi$.
This design is not applicable for face classification CNN as the number of identities $K$ are usually prohibitively large (usually in the order of ten thousands or above).

Here, to make the feature discriminative, we only learn \textbf{one} set of filters and connect the representation $\mathbf{f}(\mathbf{I})$ directly to a $K$-way softmax classification:
\begin{equation}
    \mathcal{L}_{\text{id}} =  -\log (P_{c}( \mathbf{f}(\mathbf{I}))).
\end{equation}
Here we minimize the negative log-likelihood of feature $\mathbf{f}(\mathbf{I})$ being classified to its ground-truth identity $c$.

Furthermore, global max-pooling could lead to unsatisfied recognition performance, as shown in~\cite{novotny2017anchornet} where they observed minor performance degradation compared to the model without diversity loss. One empirical explanation of this performance degradation is that max-pooling has similar effect to ReLU activation which makes the response distribution biased to the non-negative range $[0, +\infty)$. Hence it significantly limits the feasible learning space. 

Most recent works choose to use global average pooling~\cite{yi2014learning,tran2017disentangled}.
However, when applying average-pooling to introduce interpretability, it does not promote desired spatially peaky distribution. Empirically, we found the learned feature response maps of average pooling failed to have strong activation in small local regions.

Here we aim to design a pooling operation that satisfies two objectives: i) promote peaky distribution to be well-cooperated with the spatial activation diversity loss; ii) maintain the statistics of the feature responses for the global average-pooling to achieve good recognition performance. Based on these considerations, we propose the operation termed \textbf{Large magnitude filtering} (LMF), as follows:

For each channel in the feature response map, we assign $d\%$ of elements with the smallest magnitude to $0$. 
The size of the output remains the same. 
We apply $\mathcal{L}_{\text{SAD}}^{\text{response}}$ loss to the modified response map $\psi'(\mathbf{I}) \doteq g_\sigma * (\text{LMF} ( \psi(\mathbf{I})))$ in place of $\psi(\mathbf{I})$ in Eqn.~\ref{eq:div2}.
Then, the conventional global average pooling is applied to $\text{LMF} ( \psi(\mathbf{I}))$ to obtain the final representation $\mathbf{f}(\mathbf{I})$.

By removing small magnitude values from $\psi_i$, $\mathbf{f}$ won't be affected much after global average pooling, which favors discriminative feature learning. On the other hand, the peaks of the response maps are still well maintained, which leads to more reliable computation of the diversity loss.

\subsection{Feature Activation Diversity Loss}
\label{sec:fad}
One way to evaluate the effectiveness of the diversity loss is to compute the average location of the peaks within the $k$th response maps $\psi'_i(\mathbf{I})$ for an image set.
If the average locations across $K$ filters spread all over the face spatially, the diversity loss is well functioning and can associate each filer with a specific face area. 
With the SAD loss, we do observe the improved {\it spreadness} compared to the base CNN model trained without the SAD loss. 
Since we believe that more spreadness indicates  {\it higher} interpretability, we hope to further boost the spreadness of the average peak locations across filters, i.e., elements of the learnt representation. 

Motivated by the goal of learning part-based face representations, it is desirable to encourage that any local face area only affects a small subset of the filter responses. 
To fulfill this desire, we propose to create synthetic occlusion on local areas of a face image, and constrain on the difference between its feature response and that of the unoccluded original image. 
The second motivation for our proposal is to design an occlusion-robust face recognition algorithm, which, in our view, should be a natural by-product or benefit of the part-based face representation.

With this in mind, we propose a Feature Activation Diversity (FAD) Loss to encourage the network to learn filters robust to occlusions. 
That is, occlusion in a local region should only affect a small subset of elements within the representation. 
Specifically, leveraging pairs of face images $\mathbf{I}$, $\hat{\mathbf{I}}$, where $\hat{\mathbf{I}}$ is a version of $\mathbf{I}$ with a synthetically occluded region, we enforce the majority of two feature representations, $\mathbf{f}(\mathbf{I})$ and $\mathbf{f}(\hat{\mathbf{I}})$, to be similar:
\begin{equation}
\mathcal{L}_{\text{FAD}} (\mathbf{I},\hat{\mathbf{I}}) = \sum_{i }
  \left|
  \tau_i(\mathbf{I},\hat{\mathbf{I}})  \left( \mathbf{f}_i(\mathbf{I}) - \mathbf{f}_i(\hat{\mathbf{I}}) \right)
  \right|,
\label{eq:divfad}  \eqnvspace
\end{equation}
where the feature selection mask $\mathbf{\tau}(\mathbf{I},\hat{\mathbf{I}})$ is defined with threshold $t$: $\mathbf{\tau}_i(\mathbf{I},\hat{\mathbf{I}})=1$ if $\left| \mathbf{f}_i(\mathbf{I}) - \mathbf{f}_i(\hat{\mathbf{I}}) \right| < t$, otherwise $\mathbf{\tau}_i(\mathbf{I},\hat{\mathbf{I}})=0$.
There are multiple design choices for the threshold: number of elements based or value based. 
We evaluate and discuss these choices in the experiments. 

\begin{table}
\caption{\small The structures of our network architecture.} 
\label{tab:network}
\begin{center}
\small
\vspace{1mm}
\begin{tabular}{c}

\resizebox{0.8\linewidth}{!}
{
\setlength{\tabcolsep}{3pt}
\begin{tabular}{ lcccccccc }
\hline
Layer & Input & Filter/Stride & Output Size  \\ \hline \hline

conv11 & Image  & $3\times3/1$ & $96\times96\times32$  \\
conv12 & conv11 & $3\times3/1$ & $96\times96\times64$  \\
\hline
conv21 & conv12 & $3\times3/2$ & $48\times48\times64$  \\ 
conv22 & conv21 & $3\times3/1$ & $48\times48\times64$  \\
conv23 & conv22 & $3\times3/1$ & $48\times48\times128$ \\
\hline
conv31 & conv23 & $3\times3/2$ & $24\times24\times128$ \\ 
conv32 & conv32 & $3\times3/1$ & $24\times24\times96$  \\
conv33 & conv32 & $3\times3/1$ & $24\times24\times192$ \\
\hline
conv41 & conv33 & $3\times3/2$ & $12\times12\times192$ \\ 
conv42 & conv41 & $3\times3/1$ & $12\times12\times128$ \\
conv43 & conv42 & $3\times3/1$ & $12\times12\times256$ \\
\hline

conv51 & conv43 & $3\times3/2$ & $6\times6\times256$ \\ 
conv52 & conv51 & $3\times3/1$ & $6\times6\times160$ \\
conv53 & conv52 & $3\times3/1$ & $6\times6\times K$ \\
\hline

conv43-U & conv43 & upsampling  & $24\times24\times256$ \\
conv44 & conv43-U & $1\times1/1$ & $24\times24\times192$ \\
conv53-U & conv53 & upsampling  & $24\times24\times320$ \\
conv54 & conv53-U & $1\times1/1$ & $24\times24\times192$ \\
\hline
$\Phi$ (HC) & \multicolumn{1}{c}{\small conv33,44,54} & $3\times3/1$ & $24\times24\times576$ \\
$\Psi$ & $\Phi$ & $3\times3/1$ & $24\times24\times K$ & \\

AvgPool& $\Psi$ & $24\times24/1$ & $1\times1\times K$ \\ \hline
\end{tabular}}

\end{tabular}

\end{center}\figvspace \figvspace
\end{table}

We also would like to correctly classify occluded images using just subset of feature elements, which is insensitive to occlusion. 
Hence, the softmax identity loss in the occlusion branch is applied to the masked feature:
\begin{equation}
    \mathcal{L}^{\text{occluded}}_{\text{id}} =  -\log (P_{c}( \tau(\mathbf{I},\hat{\mathbf{I}}) \odot  \mathbf{f}(\hat{\mathbf{I}}))).
\end{equation}
By sharing the classifier's weights between two branches, this classifier is learned to be more robust to occlusion. 
It also leads to a better representation as filters respond to non-occluded parts need to be more discriminative.

\subsection{Implementation Details}

Our proposed method is model agnostic. 
To demonstrate this, we apply the SAD and FAD losses to two popular network architectures: one inspired by the widely used CASIA-Net~\cite{yi2014learning,tran2017representation}, the other based on ResNet50~\cite{he2016deep}. 
Tab.~\ref{tab:network} shows the structure of the former.
We add HC-descriptor-related blocks for our SAD loss learning. 
Conv33, conv44, conv54 layers are used to construct the HC descriptor via conv upsampling layers. 
We set the feature dimension $ K = 320 $. 
For ResNet50, we take the modified version in~\cite{arcface}, where $ K = 512 $. 
We also construct the HC descriptor using $3$ layers at different resolutions. 
To speed up training, we reuse the pretrained feature extraction networks shared by ~\cite{tran2017representation} and ~\cite{arcface}. 
All new weights are randomly initialized using a truncated normal distribution with std of $0.02$. 
The network is trained via SGD at an initial learning rate $10^{-3}$ and momentum $0.9$. 
The learning rate is divided by $10$ for twice when the training loss is stabled. 
We set the \textbf{LMF} rate ($ d\% $) to $ 95.83\%=1{-}\frac{24}{24{\times}24}$, i.e., keeping $24$ elements in a $24\times24$ feature response map. $g_\sigma$ is a Gaussian kernel with $\sigma=1.5$.

For FAD, the feature mask $\mathbf{\tau}$ can be computed per image pair $\mathbf{I}$ and $\hat{\mathbf{I}}$. 
However, to obtain a more reliable mask, we opt to compute $\mathbf{\tau}$ using multiple image pairs sharing the semantically equivalent occluded mask, i.e., $\mathbf{\tau}_i(\{ \mathbf{I},\hat{\mathbf{I}}\}_{n=1}^N )=1$ if $\frac{1}{N} \sum_{n=1}^N \left| \mathbf{f}_i(\mathbf{I}_n) - \mathbf{f}_i(\hat{\mathbf{I}}_n)  \right| < t$, otherwise $0$.   

\begin{figure}[t!]
\center
\includegraphics[width=1.0\linewidth]{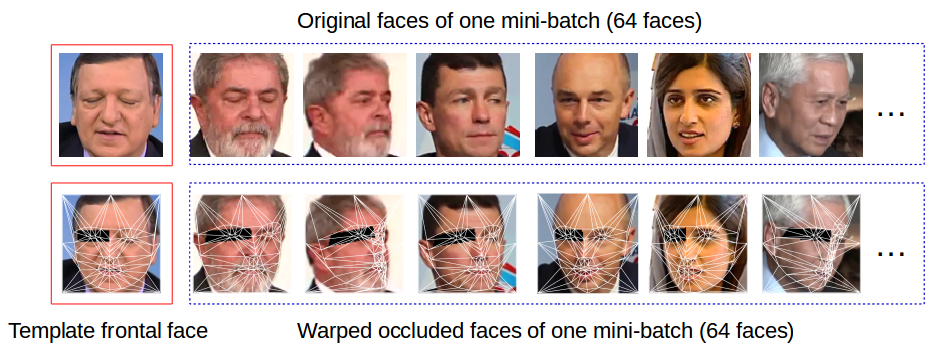}
\caption{\small 
With barycentric coordinates, we warp the vertices of the template face mask to each image within the $64$-image mini-batch.}
\label{fig:mask_warp}
\figvspace
\end{figure}

To mask the semantically equivalent local area of faces in a mini-batch regardless their poses, we first define a frontal face template with $142$ triangles created by $68$ landmarks. 
A $32\times 12$ rectangle, randomly placed on the face, is selected as a normalized mask. 
Each of the rectangle's four vertices can be represented by the barycentric coordinate w.r.t.~the triangle enclosing the vertex. 
For each image in a batch, corresponding four vertices of a quadrilateral can be found via the same barycentric coordinates. 
This quadrilateral denotes the location of a warped mask of that image (Fig.~\ref{fig:mask_warp}).

\section{Experimental Results}



\begin{figure}[t!]
\small{(a)} \includegraphics[trim=0 0 146 0, clip, width=0.26\linewidth]{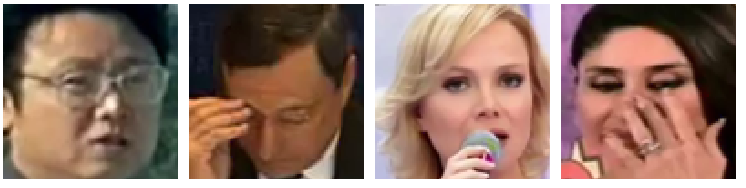}\hspace{0.7em}
\small{(b)} \includegraphics[trim=0 0 100 0, clip, width=0.26\linewidth]{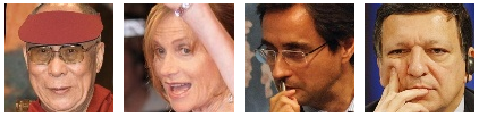}\hspace{0.7em}
\small{(c)} \includegraphics[trim=0 0 146 0, clip, width=0.26\linewidth]{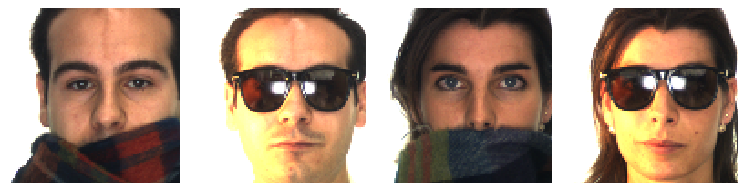}
\caption{\small Examples of (a) IJB-A, (b) IJB-C and (c) AR databases.}
\label{fig:AR_face_example}
\figvspace
\end{figure}

\Paragraph{Databases} 
We train CASIA-Net with CASIA-WebFace~\cite{yi2014learning}, ReNet50 with MS-Celeb-1M~\cite{MS-Celeb-1M}, and test on IJB-A~\cite{klare2015pushing}, IJB-C~\cite{mazepushing} and AR face~\cite{martinez1998ar} (Fig.~\ref{fig:AR_face_example}).
CASIA-WebFace contains $493,456$ images of $10,575$ subjects. 
MS-Celeb-1M includes $1$M images of $100$K subjects. 
Since it contains many labeling noise, we use a cleaned version of MS-Celeb-1M~\cite{MS-Celeb-1M}.
In our experiments, we evaluate IJB-A in three scenarios: original faces, synthetic occlusion and natural occlusion faces. 
For synthetic occlusion, we randomly generate a warped occluded area for each test image, as did in training. 
IJB-C extends IJB-A, also is a video-based face database with $3,134$ images and $117,542$ video frames of $3,531$ subjects.
One unique property of IJB-C is its label on fine-grained occlusion area. 
Thus, we use IJB-C to evaluate occlusion-robust face recognition, using test images with at least one occluded area.
AR face is another natural occlusion face database, with $\sim$$4$K faces of $126$ subjects. 
We only use AR faces with natural occlusions, including wearing glasses and scarfs. 
Following the setting in~\cite{arcface}, all training and test images are processed and resized to $112\times112$. 
Note  all ablation and qualitative evaluations use CASIA-Net, while quantitative evaluations use both models.


\subsection{Ablation Study}

\Paragraph{Different Thresholds} 
We study the affect of the threshold of FAD loss. Here we use number-of-element-based thresholding. With an abuse of notation, $t$ denotes the number of elements that the FAD loss encourages their similarity.
We train different models with $t = 130, 260, 320$.
The first three rows in Tab.~\ref{table:ijba_orignial_th} show the comparison of all three variants on IJB-A.
When forcing all elements of $\mathbf{f}(\mathbf{I}$) and $\mathbf{f}(\hat{\mathbf{I}})$ to be the same ($t=K=320$), the performance significantly drops on all three sets. 
In this case, the feature representation of the non-occluded face is negatively affected as being completely pushed toward a representation of the occluded one.
While models with $t=130$ and $260$ perform similarly, we use $t=260$ for the rest of the paper, given the observation that it makes occlusions affect less filters, pushes other filter responses away from any local occlusions, and subsequently enhances the spreadness of response locations.

\begin{table}
\small
\begin{center}
\caption{\small Ablation study on IJB-A database. 'BlaS': black mask with static sizes, 'GauD': Gaussian noise with dynamic sizes. }
\label{table:ijba_orignial_th}
\resizebox{\linewidth}{!}{
\begin{tabular}{@{\hskip .5mm}l@{\hskip 1.5mm}c@{\hskip 1.5mm}c@{\hskip 1.5mm}c@{\hskip 1.5mm}c@{\hskip 1.5mm}c@{\hskip 1.5mm}c@{\hskip .5mm}}
\toprule

Method & \multicolumn{2}{c}{IJB-A} & \multicolumn{2}{c}{Manual Occlusion} & \multicolumn{2}{c}{Natural Occlusion} \\ \cmidrule(r){1-1} \cmidrule(r){2-3} \cmidrule(r){4-5} \cmidrule(r){6-7}   

Metric ($\%$) & @FAR=$.01$ & @Rank-$1$ & @FAR=$.01$ & @Rank-$1$ & @FAR=$.01$ & @Rank-$1$ \\ \midrule

 BlaS($t=130$) & $79.0\pm1.6$ & $89.5\pm0.8$ & $76.1\pm1.7$ & $88.0\pm1.4$ & $66.2\pm4.0$ & $73.0\pm3.3$ \\

 BlaS($t=260$) & $79.2\pm1.8$ & $89.4\pm0.8$ & $76.1\pm1.4$ & $88.0\pm1.2$ & $66.5\pm6.4$ & $72.3\pm2.8$\\

 BlaS($t=320$) & $74.6\pm2.4$ & $88.9\pm1.3$ & $71.8\pm3.1$ & $87.5\pm1.6$ & $61.0\pm6.5$ & $71.6\pm3.2$\\ 
 
 GauD($t=260$) & $\mathbf{79.3}\pm2.0$ & $\mathbf{89.9}\pm1.0$ & $\mathbf{76.2}\pm2.4$ & $\mathbf{88.6}\pm1.1$ & $\mathbf{66.8}\pm3.5$ & $\mathbf{73.2}\pm3.3$\\
 
 SAD only & $78.1\pm1.8$ & $88.1\pm1.1$ & $66.6\pm5.6$ & $81.2\pm1.9$ & $64.2\pm6.9$ & $71.0\pm3.3$ \\
 
 FAD only & $76.7\pm2.0$ & $88.1\pm1.1$ & $75.2\pm2.4$ & $85.1\pm1.2$ & $66.5\pm6.4$ & $72.3\pm2.8$ \\

 \bottomrule

\end{tabular}
}
\end{center}
\figvspace
\vspace{-4mm}
\end{table}

\Paragraph{Different Occlusions and Dynamic Window Size} In FAD loss, we use the warped black window as the synthetic occlusion. 
It is important to introduce another type of occlusion to see its effects on face recognition. 
Thus, we use Gaussian noise to replace the black color in the window. 
Further, we employ a dynamic window size by randomly generating a value from $\left [ 12, 32 \right ]$ for both the window height and width. 
The face recognition results on IJB-A are shown in Tab.~\ref{table:ijba_orignial_th}, where 'BlaS' means black window with static sizes, while 'GauD' means Gaussian noise window with dynamic sizes. 
It is interesting to find that the performance of 'GauD' is slightly better. Comparing to black window, Gaussian noise contains more diverse adversarial cues.

\Paragraph{Spatial vs.~Feature Diversity Loss} Since we propose two different diversity losses, it is important to evaluate their respective effects on face recognition. 
As in Tab.~\ref{table:ijba_orignial_th}, we train our models using either loss, or both of them. 
We observe that, while the SAD loss performs reasonably well on general IJB-A, it suffers from data with occlusions, being synthetic or natural.
Alternatively, using only the FAD loss can improve the performance on the two occlusion datasets.
Finally, using both losses, the row of `BlaS($t=260$)', improves upon both models with only one loss.

\subsection{Qualitative Evaluation}

\Paragraph{Spreadness of Average Locations of Filter Response} Given an input face, our model computes $\text{LMF}(\psi(\mathbf{I}))$, the $320$ feature maps of size $ 24 \times 24 $, where the average pooling of one map is one element of the final $320$-d feature representation. 
Each feature map contains both the positive and negative response values, which are distributed at different spatial areas of the face. 
We select the locations of both the highest value for positive response and the lowest value for negative response as the {\it peak response locations}. 
To illustrate the spatial distribution of peak locations, we randomly select $1,000$ test images and calculate the weighted average location for each filter, with three notes.
1) there are two types of locations, for the highest (positive) and lowest (negative) responses respectively.
2) since the filters are responsive to semantic facial components, their $2$D spatial locations may vary with pose. 
To compensate that, we warp the peak location in an arbitrary-view face to a canonical frontal-view face, by its barycentric coordinates w.r.t.~the triangle enclosing it. 
Similar to Fig.~\ref{fig:mask_warp}, we use $68$ estimated landmarks~\cite{dense-face-alignment,jourabloo2017pose} and control points on the image boundary to define the triangular mesh.
3) the weight of each image is determined by the magnitude of its peak response.

\begin{figure}[t!]
    \centering
    \begin{tabular}{@{}l@{\hskip 0.5mm}l@{\hskip 2mm}l@{\hskip .5mm}l@{\hskip 2mm}l@{\hskip .5mm}l@{}}
&
    \includegraphics[trim = 0mm 0mm 130mm 0mm, clip, height=1.95cm]{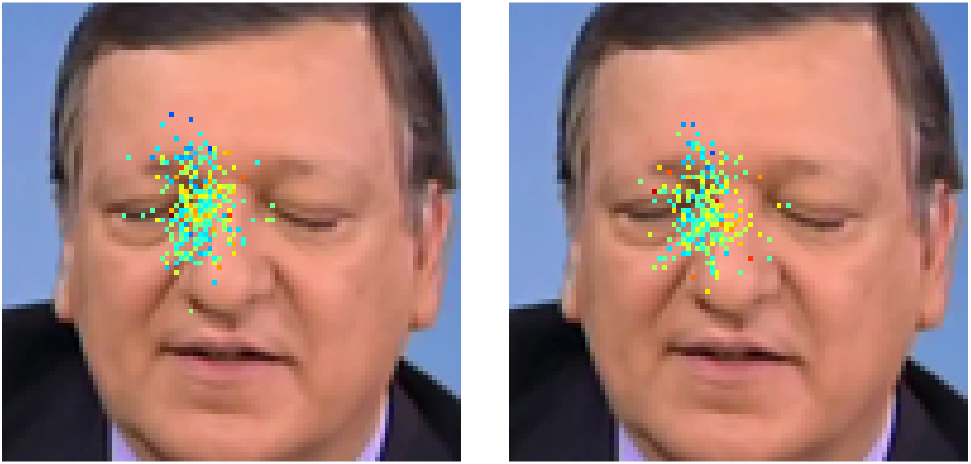} &
&
    \includegraphics[trim = 0mm 0mm 130mm 0mm, clip, height=1.95cm]{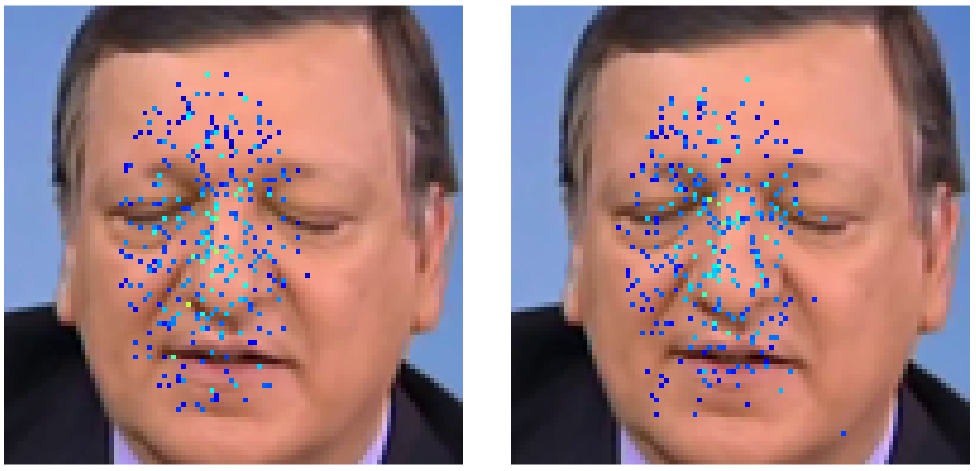} &
&
    \includegraphics[trim = 0mm 0mm 145mm 0mm, clip, height=1.95cm]{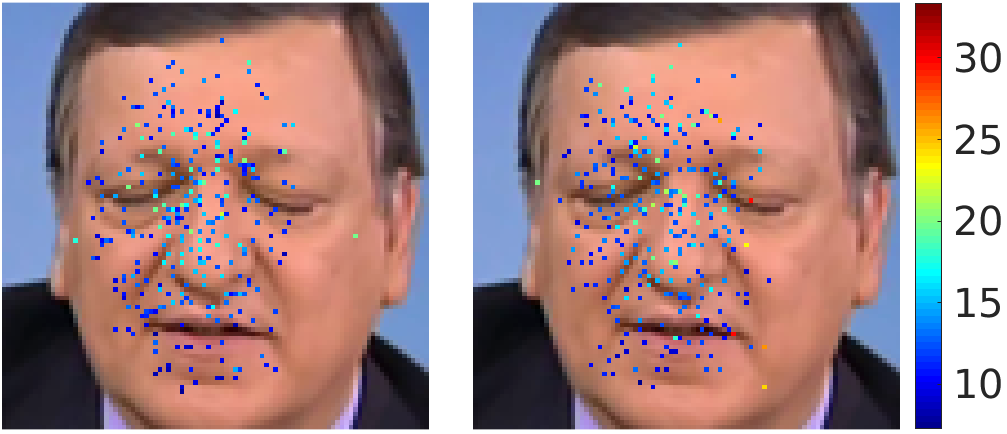}\vspace{-1mm}  \\ 
\small{(a)}&
    \includegraphics[trim = 133mm 0mm 0mm 0mm, clip, height=1.95cm]{img/avg_loc_base_new.png} &
\small{(b)}&
    \includegraphics[trim = 133mm 0mm 0mm 0mm, clip, height=1.95cm]{img/avg_loc_spatial_new.png} &
\small{(c)}&
    \includegraphics[trim = 126mm 0mm 0mm 0mm, clip, height=1.95cm]{img/avg_loc_both_new.png} \\
    \end{tabular}
\vspace{1mm} 
    \caption{\small The average locations of $320$ filters' peak responses (top: positive, bottom: negative responses) for three models: (a) base CNN ($\bar{d}=6.9$), (b) our (SAD only,  $\bar{d}=17.1$), and (c) our model ($\bar{d}=18.7$), where $\bar{d}$ quantifies the average locations spreadness. The color on each location denotes the standard deviation of peak locations. The face image size is $96 \times 96$.}
\label{fig:avg_loc}
\figvspace \vspace{1mm} 
\end{figure}

With that, the average locations for all feature maps are shown in Fig.~\ref{fig:avg_loc}. 
To compare the visualization results between our models and CNN base model,  we compute $\bar{d} = \frac{1}{K}\sum_{i}^{K}\left \| c_i-\frac{1}{K}\sum_{i}^{K}c_i \right \|$ to quantify the average locations spreadness, where $c_i$ denotes the $(x,y)$ coordinates of the $i$th average location. 
For both the positive and negative peak response, we take the mean of their $\bar{d}$. 
As in Fig.~\ref{fig:avg_loc}, our model with SAD loss enlarges the spreadness of average locations. 
Further, our model with both losses continues to push the filter responses apart from each other. 
This demonstrates that indeed our model is able to push filters to attach to diverse face areas, while in the base model all filters don't attach to specific facial part, results in average locations near the image center (Fig.~\ref{fig:avg_loc} (a)). 
In addition, we compute the standard deviation for each filter's peak location. %
With much smaller standard deviations, our model can better concentrate on a local part than the base model.

In above analysis, we set the LMF rate $d$ to $ 95.83\% $. 
It is worthy to ablate the impact of the rate $d$. 
We train models with $d=0\% $, $ 75\% $, $ 87.5\% $ or $ 95.83\% $. 
Since before average pooling the feature map is of $ 24 \times 24 $, the last $ 3 $ choices mean that we remove $ 24\times 18$, $ 24\times21 $ and $ 24\times23 $ responses respectively and $ 0\% $ denotes the base model. 
Tab.~\ref{table:std_dist} compares the average of standard deviations of peak locations across $320$ filters.
Note the values of the best model ($12.9/13.4$) equals to the average color of Fig.~\ref{fig:avg_loc}(c).
When using a larger LMF rate, the model tends to be more concentrated onto a local facial part. 
For this reason, we set $d=95.83\%$.

\begin{table}[t!]
\footnotesize
\begin{center}
\caption{\small Compare standard deviations of peaks with varying $d$.}
\label{table:std_dist}
\begin{tabular}{@{\hskip .5mm}l@{\hskip 1.5mm}c@{\hskip 1.5mm}c@{\hskip 1.5mm}c@{\hskip 1.5mm}c@{\hskip .5mm}}

\toprule
LMF ($d\%$)  & $0$ 	&	 $75.00$ 	& 	$87.50$		& 	$95.83$ \\ \midrule
std (pos./neg.) & $ 25.7/25.7 $ & $14.7/14.4$ & $ 13.5/14.0 $ & $ 12.9/13.4 $ \\ \bottomrule

\end{tabular}
\end{center}
\figvspace
\vspace{-4mm}
\end{table}

\begin{figure}[t!]
\centering \vspace{-2mm}
\begin{tabular}{@{}c@{}c@{}c@{}c@{}}
     \small{(a)}  &
    \includegraphics[width=0.48\linewidth]{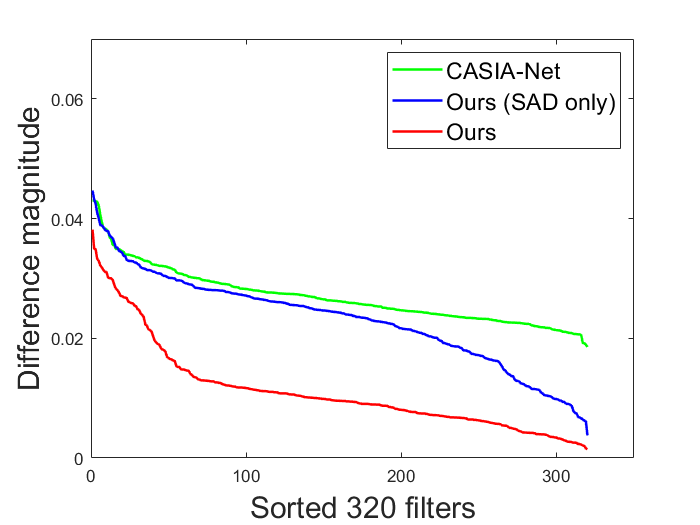} &
    \small{(b)}  &
    \includegraphics[width=0.48\linewidth]{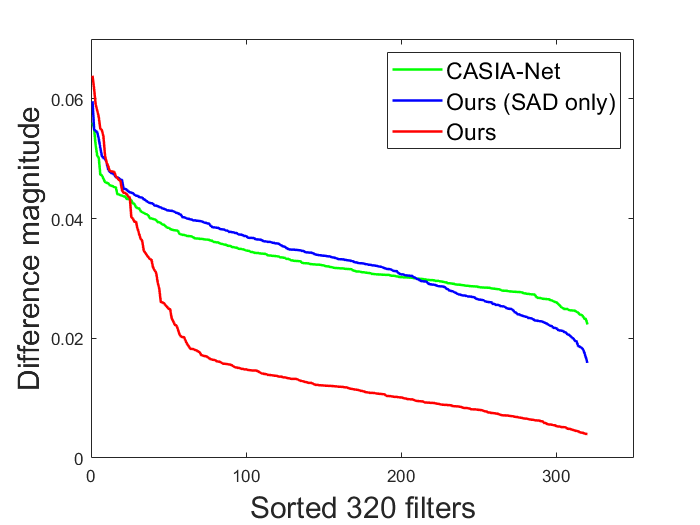}  \\
    \end{tabular}
\caption{\small Mean of feature difference on two occluded parts: (a) eye part, and (b) nose part.}
\label{fig:feat_diff}.
\figvspace\vspace{-2mm}
\end{figure}


\Paragraph{Mean Feature Difference Comparison} Both of our losses promotes part-based feature learning, which leads to occlusion robustness. Especially, in FAD, we directly minimize the difference in a portion of representation of faces with and without occlusion. 
We now study the effect of our loss on faces with occlusion.
Firstly, we randomly select $N=1,000$ test faces in different poses and generate the synthetic occlusion. After that, for each filter, we calculate the mean of feature difference between the original and occluded faces $( \frac{1}{N} \sum_{n=1}^{N} | \mathbf{f}_i(\mathbf{I}_n) - \mathbf{f}_i(\hat{\mathbf{I}}_n) | )$ for $i=1,2,\dots,K$. 
Fig.~\ref{fig:feat_diff} (a) and (b) illustrates the sorted feature difference of three models at two different occlusion parts, eye and nose, respectively.
Compare to the base CNN (trained with $\mathcal{L}_{id}$), both of our losses have smaller magnitude of differences. 
Diversity properties of SAD loss could help to reduce the feature change on occlusion, even without directly minimizing this difference. 
FDA loss further enhances robustness by only letting the occlusion modify a small portion of the representation, keeping the remaining elements invariant to the occluded part.

\begin{figure}[t!]

\centering
\includegraphics[width=0.19\linewidth]{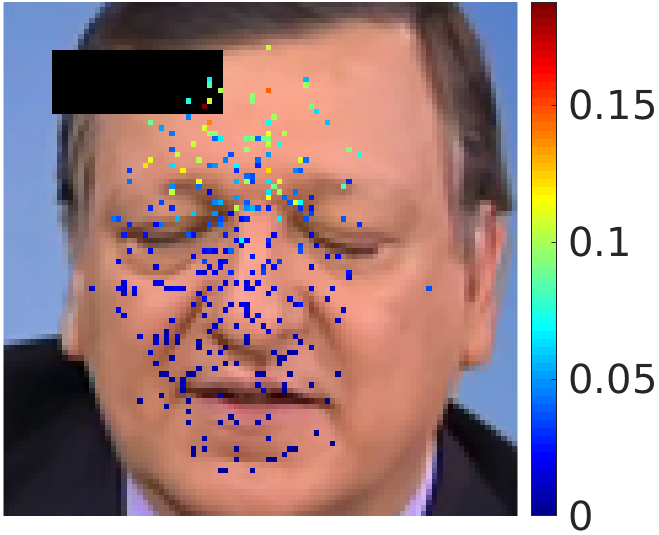}
\includegraphics[width=0.19\linewidth]{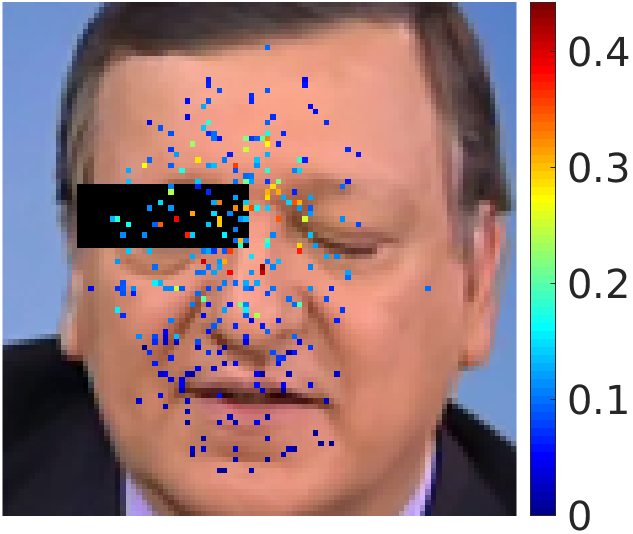}
\includegraphics[width=0.19\linewidth]{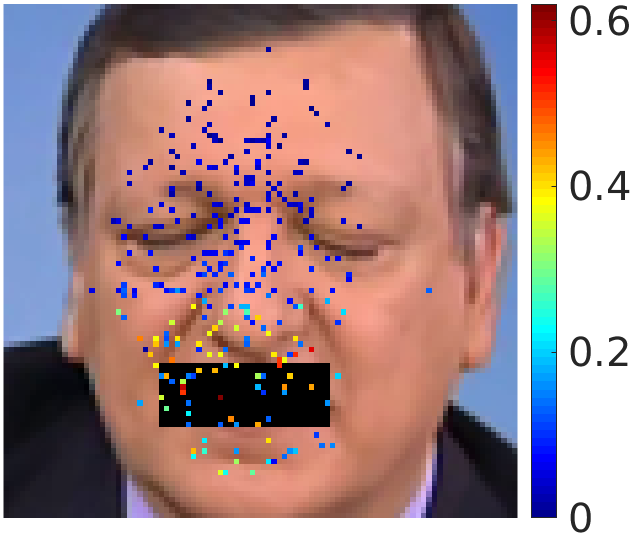}
\includegraphics[width=0.19\linewidth]{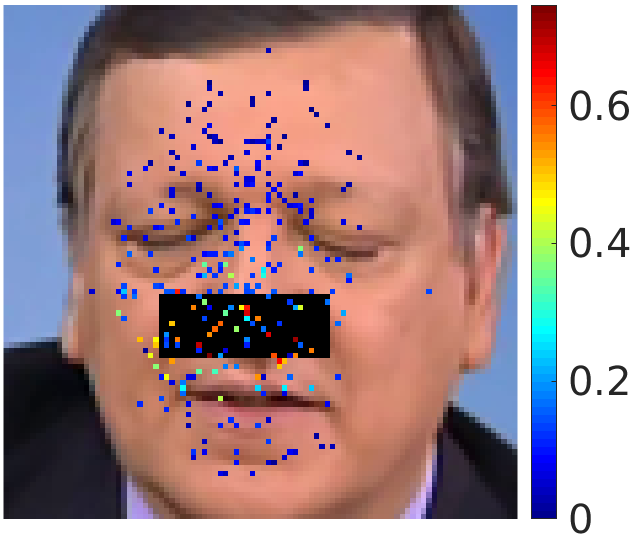}
\includegraphics[width=0.19\linewidth]{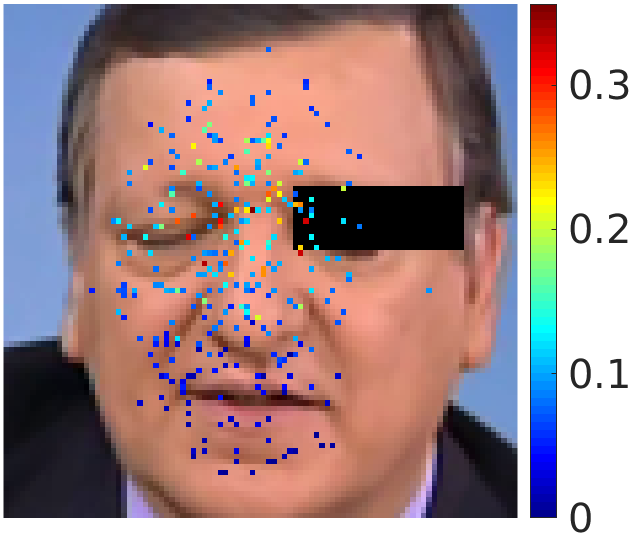}
\caption{\small The correspondence between feature difference magnitude and occlusion locations. Best viewed electronically.}
\label{fig:feat_diff_vector}
\figvspace
\end{figure}

\Paragraph{Visualization on Feature Difference Vectors} Fig.~\ref{fig:avg_loc} demonstrates that each of our filter spatially corresponds to a face location. Here we further study the relation of these average locations and semantic meaning on input images.
In Fig.~\ref{fig:feat_diff_vector}, we visualize the magnitude of each feature difference due to five different occlusions. 
We observe the locations of points with large feature difference are around the occluded face area, which means our learned filters are indeed sensitive to various facial areas. 
Further, the magnitude of the feature difference can vary with different occlusions. 
The maximum feature difference can be as high as $0.6$ with occlusion in eye or mouth, meanwhile this number is only $0.15$ in less critical area, e.g., forehead.

\begin{figure}[t!]
\centering
 \includegraphics[width=0.99\linewidth]{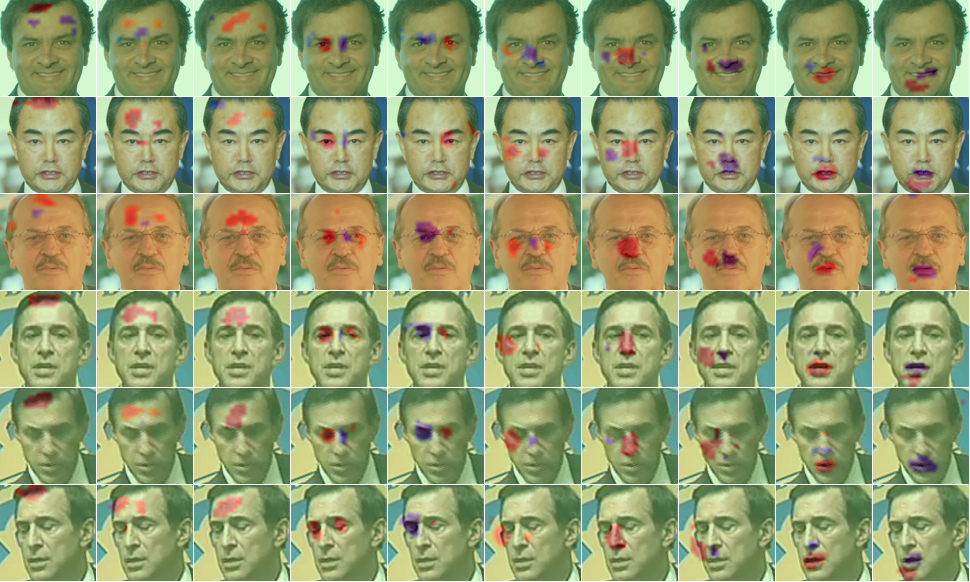}
\vspace{1mm}
\caption{\small Visualization of filter response ``heat maps" of $10$ different filters on faces from different subjects (top $3$ rows) and the same subject (bottom $3$ rows). The positive and negative responses are shown as two colors within each image. Note the high consistency of response locations across subjects and across poses.}
\label{fig:feat_visual_sorted}
\figvspace
\end{figure}

\Paragraph{Filter Response Visualization} Fig.~\ref{fig:feat_visual_sorted} visualizes the feature responses of some filters on different subjects' faces. 
From the heat maps, we can see how each filter is attached to a specific {\it semantic} location on the faces, independent to either identities or poses. 
This is especially impressive for faces with varying poses, in that despite no pose prior is used in training, the filter can always respond to the {\it semantically equivalent} local part. 

\subsection{Quantitative Evaluation on Benchmarks}

Our main objective is to show how we can improve the interpretability of face recognition while maintaining the recognition performance. Hence, the main comparison is between our proposed method and the base CNN model with the conventional softmax loss. 
Also, to show that our method is model agnostic, we use two different base CNN models, CASIA-Net and ResNet50.
Our proposed method and the respective base model only differs in the loss functions.
E.g., both our CASIA-Net-based model and base CASIA-Net model use the same network architecture as Tab.~\ref{tab:network}. 
Also, we perform data augmentation where the same synthetic faces that trained our models are fed to the training of base CASIA-Net model.
We test on two types of datasets: the generic in-the-wild faces and occlusion faces.

\begin{table*}[t!]
\footnotesize
\begin{center}
\caption{\small Comparison on three databases with occlusions.}
\label{table:ijba_manual}
\vspace{-2mm}
\resizebox{\textwidth}{!}{
\begin{tabular}{@{\hskip .5mm}l@{\hskip 1.5mm}c@{\hskip 1.5mm}c@{\hskip 1.5mm}c@{\hskip 1.5mm}c@{\hskip 1.5mm}c@{\hskip 1.5mm}c@{\hskip 1.5mm}c@{\hskip 1.5mm}c@{\hskip 1.5mm}c@{\hskip 1.5mm}c@{\hskip 1.5mm}c@{\hskip 1.5mm}c@{\hskip 1.5mm}c@{\hskip 1.5mm}c@{\hskip 1.5mm}c@{\hskip 1.5mm}c}
\toprule
Dataset 
&  \multicolumn{4}{c}{IJB-A synthetic occlusion}
&& \multicolumn{4}{c}{IJB-A natural occlusion}
&&  \multicolumn{4}{c}{IJB-C natural occlusion}
\\
\cmidrule(r){1-1} 
\cmidrule(r){2-5} 
\cmidrule(r){7-10}
\cmidrule(r){12-15}
Method $\downarrow$
&  \multicolumn{2}{c}{Verification} & \multicolumn{2}{c}{Identification} 
&& \multicolumn{2}{c}{Verification} & \multicolumn{2}{c}{Identification}
&& \multicolumn{2}{c}{Verification} & \multicolumn{2}{c}{Identification} 
\\
\cmidrule(r){1-1} 
\cmidrule(r){2-3} \cmidrule(r){4-5} 
\cmidrule(r){7-8} \cmidrule(r){9-10}
\cmidrule(r){12-13} \cmidrule(r){14-15}
Metric ($\%$) $\to$ 
&  @FAR=$.01$ & @FAR=$.001$ & @Rank-$1$ & @Rank-$5$
&& @FAR=$.01$ & @FAR=$.001$ & @Rank-$1$ & @Rank-$5$
&& @FAR=$.01$ & @FAR=$.001$ & @Rank-$1$ & @Rank-$5$ \\ \midrule

DR-GAN \cite{tran2017representation} & $61.9\pm4.7$ & $35.8\pm4.3$ & $80.0\pm1.1$ & $91.4\pm0.8$
&&
$64.7\pm4.1$ & $41.8\pm6.4$ & $70.8\pm3.6$ & $81.7\pm2.9$
&&
$82.4$ & $66.1$ & $70.8$ & $82.8$ 
\\
CASIA-Net & $61.8\pm5.5$ & $39.1\pm7.8$ & $79.6\pm2.1$ & $91.4\pm1.2$
&&
$64.4\pm6.1$ & $40.7\pm6.8$ & $71.3\pm3.5$ & $81.6\pm2.5$
&&
$83.3$ & $67.0$ & $72.1$ & $83.3$
\\
Ours (CASIA-Net) & $76.2\pm2.4$ & $55.5\pm5.7$ & $88.6\pm1.1$ & $95.0\pm0.7$ 
&&
$66.8\pm3.4$ & $48.3\pm5.5$ & $73.2\pm2.5$ & $82.3\pm3.3$
&&
$83.8$ & $69.3$ & $74.5$ & $83.6$
\\ \hline

ResNet50 \cite{he2016deep} & $93.0\pm0.7$ & $80.9\pm4.7$ & $92.8\pm0.9$ & $95.5\pm0.8$
&&
$\mathbf{86.0}\pm1.8$ & $64.3\pm7.7$ & $79.8\pm4.2$ & $84.9\pm3.1$ 
&&
$93.1$ & $89.0$ & $\mathbf{87.5}$ & $\mathbf{91.0}$ 
\\
Ours (ResNet50)& $\mathbf{94.2}\pm0.6$ & $\mathbf{87.5}\pm1.5$ & $\mathbf{93.4}\pm0.7$ & $\mathbf{95.8}\pm0.4$
&&
$\mathbf{86.0}\pm1.6$ & $\mathbf{72.6}\pm5.0$ & $\mathbf{80.0}\pm3.2$ & $\mathbf{85.0}\pm3.1$
&&
$ \mathbf{93.4} $ & $ \mathbf{89.8} $ & $ 87.4 $ & $ 90.7 $ 
 \\ \bottomrule
\end{tabular}}
\end{center}
\figvspace
\vspace{-3mm}
\end{table*}

\begin{table}[t!]
\footnotesize
\begin{center}
\caption{\small Comparison on IJB-A database.}
\label{table:ijba_orignial}
\vspace{-2mm}
\begin{tabular}{@{\hskip .5mm}l@{\hskip 1.5mm}c@{\hskip 1.5mm}c@{\hskip 1.5mm}c@{\hskip 1.5mm}c@{\hskip .5mm}}

\toprule
Method $\downarrow$ & \multicolumn{2}{c}{Verification} & \multicolumn{2}{c}{Identification} \\ \cmidrule(r){1-1} \cmidrule(r){2-3} \cmidrule(r){4-5}
Metric ($\%$) $\to$ & @FAR=$.01$ & @FAR=$.001$ & @Rank-$1$ & @Rank-$5$ \\ \midrule

DR-GAN \cite{tran2017representation} & $79.9\pm1.6$ & $56.2\pm7.2$ & $88.7\pm1.1$ & $95.0\pm0.8$ \\
CASIA-Net & $74.3\pm2.8$ & $49.0\pm7.4$ & $86.6\pm2.0$ & $94.2\pm0.9$ \\
CASIA-Net agu. & $78.9\pm1.8$ & $56.6\pm4.8$ & $88.5\pm1.1$ & $94.9\pm0.8$ \\
Ours (CASIA-Net) & $79.3\pm2.0$ & $60.2\pm5.5$ & $89.9\pm1.0$ & $95.6\pm0.6$ \\ \hline
FaceID-GAN \cite{faceIDGAN} & $ 87.6\pm1.1 $ & $ 69.2\pm2.7 $ & $ - $ & $ - $ \\
VGGFace2 \cite{vggface2} & $93.9\pm1.3$ & $85.1\pm3.0$ & $\mathbf{96.1}\pm0.6$ & $\mathbf{98.2}\pm0.4$ \\
PRFace\cite{PRfaceRecog} & $ 94.4\pm0.9 $ & $ 86.8\pm1.5 $ & $ 92.4\pm1.6 $ & $ 96.2\pm1.0 $ \\ 
ResNet50 \cite{he2016deep} & $\mathbf{94.8}\pm0.6$ & $86.0\pm2.6$ & $94.1\pm0.8$ & $96.1\pm0.6$ \\
Ours (ResNet50)& $94.6\pm0.8$ & $\mathbf{87.9}\pm1.0$ & $93.7\pm0.9$ & $96.0\pm0.5$ \\ \bottomrule
\end{tabular}
\end{center}
\figvspace
\end{table}

\begin{table}[t!]
\footnotesize
\begin{center}
\caption{\small Comparison on IJB-C database.}
\label{table:ijbc_natural}
\begin{tabular}{@{\hskip .5mm}l@{\hskip 1.5mm}c@{\hskip 1.5mm}c@{\hskip 1.5mm}c@{\hskip 1.5mm}c@{\hskip .5mm}}
\toprule
Method $\downarrow$ & \multicolumn{2}{c}{Verification} & \multicolumn{2}{c}{Identification} \\ \cmidrule(r){1-1} \cmidrule(r){2-3} \cmidrule(r){4-5}
Metric ($\%$) $\to$ & @FAR=$.01$ & @FAR=$.001$ & @Rank-$1$ & @Rank-$5$ \\ \midrule

 DR-GAN \cite{tran2017representation} & $88.2$ & $73.6$ & $ 74.0 $ & $ 84.2 $ \\ 

 CASIA-Net & $87.1$ & $72.9$ & $74.1$ & $83.5$ \\ 
 
 
 Ours (CASIA-Net) & $89.2$ & $75.6$ & $77.6$ & $86.1$ \\ \hline
 
 VGGFace2 \cite{vggface2} & $95.0$ & $90.0$ & $89.8$ & $\mathbf{93.9}$ \\
 
 Mn-v \cite{Mnv} & $ \mathbf{96.5} $ & $ 92.0 $ & $ - $ & $ - $ \\

 AIM \cite{AIM} & $ 96.2 $ & $ \mathbf{93.5} $ & $ - $ & $ - $ \\

 ResNet50 \cite{he2016deep} & $95.9$ & $93.2$ & $\mathbf{90.5}$ & $93.2$ \\ 
 
 Ours (ResNet50) & $ 95.8 $ & $ 93.2 $ & $ 90.3 $ & $ 93.2 $ \\ \bottomrule
\end{tabular}
\end{center}
\figvspace
\end{table}

\Paragraph{Generic in-the-wild faces}
As shown in Tabs.~\ref{table:ijba_orignial},~\ref{table:ijbc_natural}, when comparing to the base CASIA-Net model, our CASIA-Net-based model with two losses achieves the superior performance. 
The same superiority is demonstrated w.r.t.~CASIA-Net with data augmentation, which shows that the gain is caused by the novel loss function design.
For the deeper ResNet50 structure, our proposed model achieves similar performance as the base model, and both outperform the models with CASIA-Net as the base.
Even comparing to state-of-the-art methods, the performance of our ResNet50-based model is still competitive.
It is worthy note that this is the first time that {\it a reasonably interpretable representation is able to demonstrate competitive state-of-the-art recognition performance on a widely used benchmark}, e.g., IJB-A.


\Paragraph{Occlusion faces}
We test our models and base models on multiple occlusion face datasets.
The synthetic occlusion of IJB-A, the natural occlusion of IJB-A, and the natural occlusion of IJB-C have $500/25,795$,  $466/12,703$, and $3,329/78,522$ subjects/images, respectively.
As shown in Tab.~\ref{table:ijba_manual}, the performance improvement on the occlusion datasets are more substantial than the generic IJB-A database, which shows the advantage of interpretable representations in handling occlusions.

For AR faces, we select all $810$ images with eyeglasses and scarfs occlusions, from which $6,000$ same-person and $6,000$ different-person pairs are randomly selected.
We compute the representations of an image pair and its cosine distance. 
The Equal Error Rates of CASIA-Net, ours (CASIA-Net), ResNet50 and ours (ResNet50) are $21.6\%$,  $16.2\%$, $ 4.2\% $ and $3.9\% $, respectively.



\subsection{Other Application}

\begin{figure}
    \centering
    \begin{tabular}{@{}l@{\hskip 1.5mm}l@{}}
    \includegraphics[height=3cm]{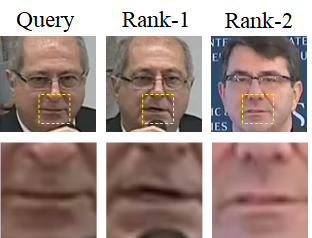} 
    &
    \includegraphics[height=3cm]{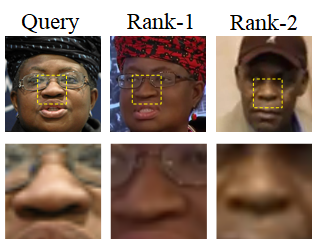} 
    \end{tabular}
    \caption{\small Partial face retrieval with mouth (left),  and nose (right).}
\label{fig:part_face_recog}
\figvspace 
\end{figure}

Despite the interpretable face recognition,  another potential application of our method is partial face retrieval. 
Assuming we are interested in retrieving images with similar noses, we can define ``nose filters" base on filters' average peak location with our models, as in Fig.~\ref{fig:avg_loc}. 
Then, the part-based feature for retrieving is constructed by masking out all elements in the identity feature $\mathbf{f}(\mathbf{I})$ except selected part-related filters. 
For demonstration,  from IJB-A test set, we select one pair of images from a subset of $150$ identities, to create a set of $300$ images in total. Using different facial parts of each image as a query, our accuracies of retrieving the remaining image of the same subject as the rank-$1$ result are $71\%$, $58\%$ and $69\%$ for eyes, mouth, and nose respectively. 
Results are visualized in Fig.~\ref{fig:part_face_recog}, we can retrieve facial parts that are not from the same identity but visually very similar to the query part.


\section{Conclusions}

In this paper, we present our efforts towards interpretable face recognition. 
Our grand goal is to learn from data a structured face representation where each dimension activates on a consistent semantic face part and captures its identity information. 
We empirically demonstrate the proposed method, enabled by our novel loss function design, can lead to more locally constrained individual filter responses and overall widely-spreading filters distribution, yet maintaining SOTA face recognition performance.
A by-product of the harnessed interpretability is improved robustness to occlusions in face recognition.

\paragraph{Acknowledgement}
This work is partially sponsored by Adobe Inc.~and Army Research Office under Grant Number W911NF-18-1-0330.  The views and conclusions contained in this document are those of the authors and should not be interpreted as representing the official policies, either expressed or implied, of the Army Research Office or the U.S. Government. The U.S. Government is authorized to reproduce and distribute reprints for Government purposes notwithstanding any copyright notation herein.

{\small
\bibliographystyle{ieee}
\bibliography{egbib}
}

\end{document}